# Simulation of Nanorobots with Artificial Intelligence and Reinforcement Learning for Advanced Cancer Cell Detection and Tracking

**Shahab Kavousinejad**
Dentofacial Deformities Research Center,
Research Institute of Dental Sciences, Shahid Beheshti University of Medical Sciences,
Tehran, Iran.
shahabkavousinejad@sbmu.ac.ir

## ABSTRACT

Nanorobots represent a transformative frontier in targeted drug delivery and the treatment of neurological disorders, with significant potential for crossing the blood-brain barrier (BBB). Leveraging advances in nanotechnology and bioengineering, these miniature devices exhibit capabilities for precise navigation and targeted payload delivery, particularly in addressing conditions like brain tumors, Alzheimer's, and Parkinson's disease. Recent developments in artificial intelligence (AI) and machine learning (ML) are enhancing the navigation and efficacy of nanorobots, enabling them to detect and interact with cancer cells through biomarker analysis. This work presents a novel reinforcement learning (RL) framework for optimizing nanorobot navigation in complex biological environments, focusing on the detection of cancer cells by analyzing the concentration gradients of surrounding biomarkers. Using a computer simulation model, we explore the behavior of nanorobots in a three-dimensional space populated with cancer cells and biological barriers. The proposed method employs Q-learning to refine movement strategies based on real-time biomarker concentration data, allowing nanorobots to autonomously navigate to cancerous tissues for targeted drug delivery. This study establishes a foundational model for subsequent laboratory experiments and clinical applications, with implications for advancing personalized medicine and developing less invasive cancer treatments. The integration of intelligent nanorobots could revolutionize therapeutic approaches, reducing side effects and improving treatment efficacy for cancer patients. Further research will explore the practical deployment of these technologies in medical settings, aiming to realize the full potential of nanorobotics in healthcare. The source code for this simulation is available on GitHub: https://github.com/SHAHAB-K93/cancer-and-smart-nanorobot

*Keywords* Nanorobots · Artificial Intelligence · Reinforcement Learning · Cancer Cell Tracking

## 1 Introduction

Nanorobots are emerging as promising tools for targeted drug delivery and the treatment of neurological disorders, particularly in crossing the blood-brain barrier (BBB) [1, 2]. These miniature devices, utilizing nano bioelectronics and biologically inspired designs, offer precise control in disease diagnosis and treatment [3]. Recent advancements in nanotechnology and microfabrication have led to the development of various micro/nanorobots (MNRs), including nano swimmers, nano engines, and biologically inspired microbats [4]. The integration of artificial intelligence (AI) and machine learning is enhancing the capabilities of nanorobots in navigation, targeting, and payload delivery across the BBB without causing damage or toxicity [1, 2]. Although promising for conditions such as brain tumors, Alzheimer's, and Parkinson's disease, a systematic investigation of nanorobots' behavior, biocompatibility, and efficacy is essential prior to clinical evaluations.

In addition to neurological applications, nanorobots are making strides in cancer diagnosis and treatment, particularly in targeting brain cancer cells [5]. These nanoscale devices, ranging from 0.1 to 10 micrometers, can navigate the bloodstream, detect cancer cells, and deliver targeted drug therapies [6]. Their ability to cross biological barriers, including the BBB, allows them to reach previously inaccessible areas [5]. Advanced designs are incorporating glucose-based cancer detectors and drug-releasing mechanisms activated by electrical signals [6]. The potential of nanorobotics extends beyond cancer treatment to various medical applications, including heart diseases, diabetes, and kidney disorders [7]. Ongoing research is focused on developing autonomous computational nanorobots for in vivo medical diagnosis and treatment [6].

Recent studies explore the application of reinforcement learning (RL) techniques, such as Q-learning, to optimize nanorobot navigation within the body. These algorithms aim to reduce the mean path length to cancer targets, thereby improving system accuracy [8]. An Improved Bacterial Foraging Optimization Algorithm (IBFOA) with cooperative learning has been proposed for multi-nanorobot cooperation in eradicating cancer cells, demonstrating enhanced performance over traditional methods [9]. Furthermore, a novel computational nano biosensing framework employs RL-based strategies for tumor targeting, combining optimal routing in the vascular network with the learning of biological gradient fields, which accelerates the search for optimal paths to tumors in complex vessel networks [10].

Cancer biomarkers play a crucial role in early detection, diagnosis, and treatment of cancer. These biomarkers can originate from tumor cells or the tumor stroma, which includes non-malignant cells and the extracellular matrix [11]. Traditional tissue biopsies for cancer diagnosis are invasive and limited. Liquid biopsies, a non-invasive alternative, analyze biomarkers in blood, including circulating tumor cells, circulating tumor DNA, and exosomes [12]. This method offers advantages such as reduced risk, cost, and diagnosis time. Cancer biomarkers encompass a wide range of biochemical entities, including nucleic acids, proteins, and metabolites[13], which can be used for risk assessment, prognosis, and predicting treatment efficacy [14]. Advancements in isolation technologies and multidisciplinary approaches are enhancing cancer biomarker detection, paving the way for personalized medicine and point-of-care diagnostics [12].

This work aims to develop a reinforcement learning framework for nanorobots that enables them to identify cancer cells by analyzing the concentration of surrounding cancer biomarkers. Initially, we present a computer simulation modeling the behavior and interactions of nanorobots within a biological context. The findings from this simulation establish a foundation for future laboratory experiments, where these algorithms can be tested in real-world scenarios. By leveraging advanced AI techniques, these smart nanorobots are designed to autonomously navigate complex biological environments, accurately detect cancer cells, and establish connections with them. Once in proximity, they can facilitate the targeted delivery of anti-cancer drugs directly into affected cells, potentially revolutionizing cancer treatment by minimizing side effects and enhancing therapeutic efficacy. The implications of this technology could significantly impact the future of personalized medicine, paving the way for more effective and less invasive treatment options for patients. Further research and development could bring the vision of intelligent nanorobots in clinical applications to fruition.

## 2 Methods

### 2.1 Biomarker Detection and Density Calculation

The movement of nanorobots in this simulation is driven by biomarker gradients. Biomarkers are biological molecules associated with tumor cells that diffuse outward into surrounding tissue. Nanorobots detect the concentration of these biomarkers and use this information to locate cancer cells. The biomarker density at any given point follows a Gaussian distribution, which serves as a realistic model for how concentration diminishes with distance from the source.

The biomarker concentration $\rho(x, y, z)$ at any position $(x, y, z)$ is modeled as:

$$\rho(x, y, z) = \rho_0 \cdot \exp\left(-\frac{(x-x_0)^2 + (y-y_0)^2 + (z-z_0)^2}{2\sigma^2}\right)$$

Where:

- $\rho_0$ is the peak concentration at the source (cancer cell location),
- $(x_0, y_0, z_0)$ denotes the coordinates of the biomarker source,
- $\sigma$ represents the spread of the biomarker concentration.

This distribution allows the nanorobots to detect the highest concentration near the source and lower concentrations as they move away. By following the gradient, nanorobots are naturally led toward cancer cells.

## 2.2 Gradient-Based Navigation

To move effectively toward cancer cells, nanorobots leverage the gradient of biomarker density in their environment. Biomarkers are biological molecules that can signal the presence of cancer cells, and their concentration can vary significantly within a given area. The nanorobots calculate the density of these biomarkers, denoted as $\rho(x, y, z)$, to determine their spatial distribution. The nanorobots use the partial derivatives of the biomarker density function with respect to spatial coordinates (x, y, z) to compute the gradients, which are represented by the following equations:

$$\frac{\partial \rho}{\partial x} = -\frac{(x - x_0)}{\sigma^2} \rho(x, y, z)$$

$$\frac{\partial \rho}{\partial y} = -\frac{(y - y_0)}{\sigma^2} \rho(x, y, z)$$

$$\frac{\partial \rho}{\partial z} = -\frac{(z - z_0)}{\sigma^2} \rho(x, y, z)$$

In these equations:

- $(x_0, y_0, z_0)$ represents the coordinates of the source of the biomarker concentration (e.g., the location of cancer cells).
- $\sigma^2$ is a parameter that reflects the spread or dispersion of the biomarker signal in space, affecting how quickly the concentration decreases with distance from the source.

The calculated gradients $\frac{\partial \rho}{\partial x}$, $\frac{\partial \rho}{\partial y}$ and $\frac{\partial \rho}{\partial z}$ indicate the rate and direction of change of the biomarker concentration in three-dimensional space. Specifically, these gradients show the direction in which the concentration of biomarkers increases most rapidly. By continuously measuring these gradients, the nanorobots can adjust their movement accordingly. If, for example, the gradient in the x-direction is positive, it indicates that moving in the positive x-direction will lead the nanorobots toward areas of higher biomarker concentration. This adaptive movement allows the nanorobots to efficiently follow the concentration slope toward the source of biomarkers, enabling them to navigate directly to cancer cells. This approach not only enhances the effectiveness of the nanorobots in locating cancer cells but also reduces the time and energy spent on navigating through the surrounding tissue. By optimizing their trajectory based on real-time biomarker concentration data, nanorobots can play a crucial role in targeted cancer therapies and diagnostic procedures, ultimately improving treatment outcomes. These gradients indicate the direction in which biomarker concentration increases most rapidly. By adjusting their position according to the calculated gradient, nanorobots efficiently follow the concentration slope toward the source, enabling direct navigation to cancer cells.

## 2.3 Reinforcement Learning for Path Optimization

While the biomarker density gradient provides directional cues, nanorobots further refine their movement strategy through reinforcement learning. Reinforcement learning algorithms allow the nanorobots to adapt to the environment by maximizing cumulative reward. In this scenario, the reward is defined by the change in biomarker concentration, encouraging nanorobots to explore paths that increase the concentration level. The nanorobot's movement decision at each step $t$ is informed by a policy function $\pi$, which dictates the probability of taking action $a_t$ based on the current state $s_t$:

$$a_t \sim \pi(s_t)$$

The objective of reinforcement learning is to maximize the expected cumulative reward $R_t$:

$$R_t = \sum_{k=0}^{\infty} \gamma^k r_{t+k}$$

where:

- $\gamma$ is the discount factor that weighs future rewards,
- $r_{t+k}$ represents the reward at step $t + k$.

By learning which actions result in increased biomarker density, nanorobots improve their path selection over time, enhancing their efficiency in reaching cancer cells.

## 2.4 Simulation Environment

This study investigates the navigation strategies of nanorobots within a three-dimensional environment, utilizing Q-learning for decision-making. The primary objective is to guide these nanorobots toward cancer cells through the detection of biomarkers while overcoming various environmental challenges. The simulation is conducted in a three-dimensional cubic space measuring 50 units per side. This environment includes several critical components:

- **Biomarkers**: Chemical substances that indicate the presence of cancer cells and serve as guides for the nanorobots.
- **Cancer Cells**: Randomly distributed throughout the space and serving as primary targets for the nanorobots.
- **Obstacles**: Represent biological barriers that nanorobots must navigate around.

Initially, nanorobots start from random positions within the environment. Their primary movement strategy involves navigating toward areas with higher concentrations of biomarkers, indicating the proximity of cancer cells. Obstacles in the simulation are generated randomly within the three-dimensional environment, mimicking the unpredictable and complex nature of real biological environments. These obstacles represent biological barriers that nanorobots might encounter, such as dense tissue, blood vessels, or cellular structures. By introducing randomly placed obstacles, the simulation reflects the variability and challenges nanorobots would face in actual physiological conditions. This randomness not only enhances the realism of the simulation but also allows for the evaluation of the nanorobots' navigation strategies under different scenarios, ensuring they can effectively maneuver around barriers while pursuing their targets, such as cancer cells.

## 2.5 Decision-Making Framework

The decision-making process of the nanorobots is modeled using Q-learning, a reinforcement learning algorithm. In this framework, each state is defined by the current position of a nanorobot, which includes:

- Nearby biomarker concentrations.
- Distances to obstacles.
- Distances to cancer cells.

Possible actions include:

- Moving toward the nearest biomarker.
- Advancing toward a cancer cell.
- Navigating around obstacles.

An epsilon-greedy strategy balances exploration and exploitation. The exploration rate (epsilon) decreases over time, allowing the nanorobots to refine their decision-making as they gather more experience.

## 2.6 Dynamic Environment Integration

To simulate realistic biological challenges, obstacles are dynamically integrated into the environment. When a nanorobot detects an obstacle within a specified radius, it recalibrates its trajectory to avoid collisions. The concentration of biomarkers is computed based on their distances from the nanorobot's current position, following the formula:

$$C(x, y) = \sum_{i=1}^{n} \frac{B_i}{d_i^2}$$

In this equation:
- $C(x, y)$ represents the biomarker concentration at the position $(x, y)$,
- $B_i$ is the concentration of the $i$-th biomarker,

- $d_i$ is the distance to the $i$-th biomarker, calculated as:

$$d_i = \sqrt{(x - x_i)^2 + (y - y_i)^2 + (z - z_i)^2}$$

where $(x_i, y_i, z_i)$, are the coordinates of the i-th biomarker. The formula indicates that the concentration of each biomarker decreases with the square of its distance from the nanorobot, which is a common assumption in models of concentration decay in three-dimensional space.

### 2.7 Learning and Adaptation

The Q-learning algorithm updates action values based on the experiences of the nanorobots. The update rule is defined as:

$$Q(s, a) \leftarrow Q(s, a) + \alpha \left[ r + \gamma \max_{a'} Q(s', a') - Q(s, a) \right]$$

Where $Q(s, a)$ is the action-value function for state $s$ and action $a$, $\alpha$ is the learning rate, $r$ is the reward received after taking action $a$ in state $s$, $\gamma$ is the discount factor, $s'$ is the new state after taking action $a$. And $\gamma \max_{a'} Q(s', a')$ is the maximum Q-value over all possible actions $a'$ in the next state of $s'$.

- **Current Q-value** ($Q(s, a)$): This is the existing estimate of the value of taking action $a$ in the current state $s$. The Q-value represents the expected cumulative reward an agent can achieve by choosing $a$ in $s$, assuming it continues to make optimal decisions thereafter.
- **Reward** ($r$): $r$ is the immediate reward the agent receives after taking action $a$ in state $s$. This reward directly influences how the Q-value is updated, encouraging the agent to repeat actions that yield high rewards.
- **Future Q-value** ($\max_{a'} Q(s', a')$): $\max_{a'} Q(s', a')$ represents the maximum estimated Q-value in the new state $s'$ (after taking action $a$ in $s$). By taking the maximum, the agent optimistically assumes that it will take the best possible future action, encouraging it to select actions that lead to states with high potential for reward.
- **Learning Rate** ($\alpha$): $\alpha$ ($0 < \alpha \leq 1$) controls the rate at which the agent updates its Q-values. A high $\alpha$ means the agent gives more weight to new information, while a low $\alpha$ means it relies more on past knowledge.
- **Discount Factor** ($\gamma$): $\gamma$ ($0 \leq \gamma \leq 1$) determines how much future rewards are valued compared to immediate rewards. A higher $\gamma$ makes the agent value long-term rewards, whereas a lower $\gamma$ makes it focus more on immediate rewards.

The agent updates $Q(s, a)$ by adding the difference between the expected reward (based on current information) and the old estimate. This update pushes $Q(s, a)$ closer to the "true" value based on the reward and future optimal Q-values. The agent repeatedly applies this formula during its interactions, gradually improving its Q-values to reflect the best possible actions in each state. Over time, the agent learns an optimal policy, maximizing its cumulative reward by choosing actions that lead to high future rewards.

The reward function is defined as follows:

$$r = \begin{cases} +10 & \text{if a cancer cell is reached} \\ -1 & \text{if an obstacle is hit} \\ +1 & \text{for every unit moved toward a biomarker} \end{cases}$$

The simulation iterates over numerous episodes, tracking the movements of nanorobots as they navigate toward cancer cells. Figure 1 illustrates the simulation of the system featuring a single cancer cell (red) and one nanorobot (blue), showcasing the direct interaction between the two entities. Figure 2 expands on this concept by simulating the environment with two cancer cells (red) and multiple nanorobots (blue). In this scenario, black spheres are depicted as obstacles that the nanorobots must skillfully navigate to successfully reach their targets. In the simulation illustrated in Figure 2, the nanorobots (blue) successfully navigate through the complex environment and reach the cancer cells (red). Q-learning is a powerful reinforcement learning algorithm that enhances the decision-making capabilities of nanorobots in dynamic and unpredictable environments. Unlike rule-based methods, which struggle to adapt to new conditions, Q-learning enables nanorobots to learn from their experiences by updating Q-values based on rewards received during interactions with their surroundings. This ability allows them to strike a balance between exploration and exploitation, optimizing their strategies over time. Additionally, Q-learning equips nanorobots with the flexibility to quickly respond to environmental changes and identify the most efficient paths toward their targets, such as cancer cells, thereby improving their overall effectiveness in complex tasks.

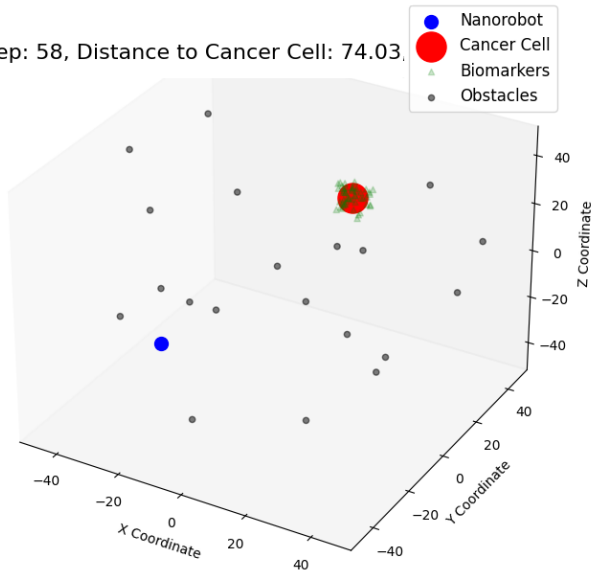
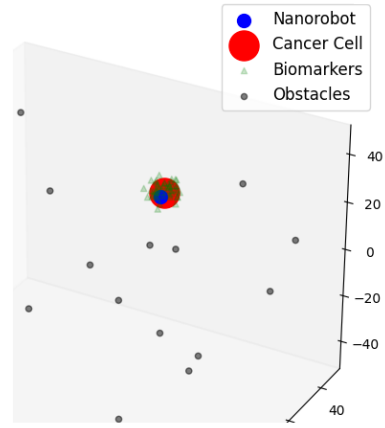

Figure 1: Simulation of the system for a single cancer cell (red) and one nanorobot (blue).

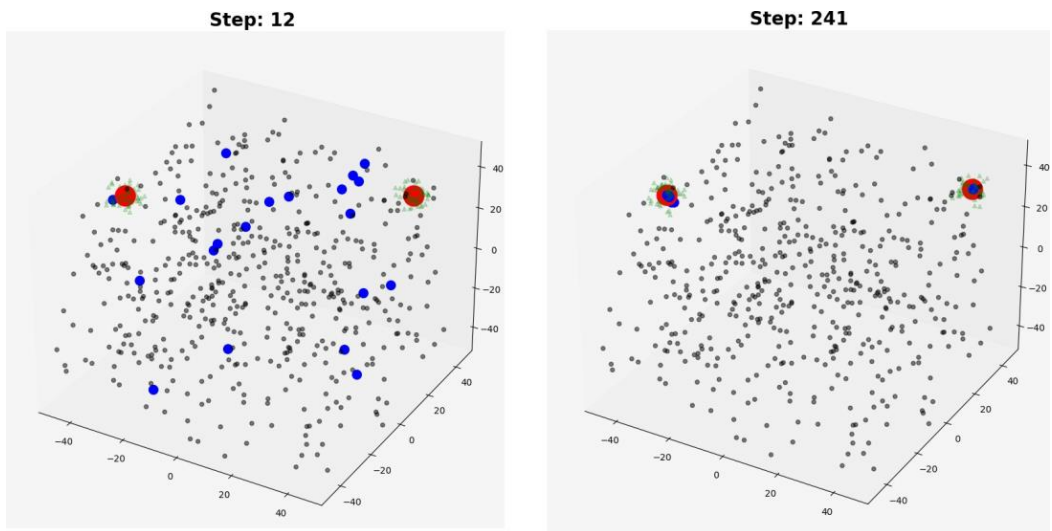

Figure 2: Simulation of the system for two cancer cells (red) and several nanorobots (blue). The black spheres represent obstacles that the nanorobots must navigate through to reach their target.

## 3 Results

The following metrics and visualizations were used to assess the nanorobot's performance, efficiency, and effectiveness in targeting the cancer cell through a challenging environment with biomarker concentrations and obstacles. Key performance indicators included distance measurements, biomarker concentration, total steps, and convergence trends. Below are the detailed descriptions of each figure, along with the metrics that drive the analysis.

### 3.1 Path Simulation and Convergence

Figure 3 displays the three-dimensional trajectory of the nanorobot as it navigates toward the cancer cell. The blue line represents the path taken by the nanorobot, illustrating its movement through the spatial environment. In contrast, the red line indicates the path of the cancer cell, which exhibits random movement within a confined space. This dynamic behavior adds complexity to the navigation task, as the nanorobot must detect the cancer cell from a distance and adapt its path accordingly to reach it. Additionally, the plot features black spheres representing obstacles encountered by the nanorobot during its journey. These obstacles are crucial for understanding the challenges the nanorobot faces as it maneuvers through a complex environment. The 3D visualization effectively showcases how the nanorobot strategically navigates around these barriers while progressing toward the cancer cell, highlighting its pathfinding capabilities in a realistic setting.

### 3.2 Biomarker Concentration and Distance Analysis

Figure 4 illustrates the relationship between biomarker concentration and the distance of the nanorobot from the target cancer cell as the nanorobot moves. As depicted in the graph, with each step the nanorobot takes, the sensed biomarker concentration (represented by the red line) increases significantly. This increase indicates that the nanorobot is successfully navigating toward areas with higher biomarker densities, which are indicative of the cancer cells. Simultaneously, the blue line shows a decrease in the distance to the cancer cell over the same steps. This trend indicates that the nanorobot is not only detecting higher concentrations of biomarkers but is also effectively reducing its distance to the target, highlighting its efficient navigation capabilities. The combined trends of increasing biomarker concentration and decreasing distance suggest that the nanorobot is on a successful trajectory toward the cancer cells, optimizing its path based on the gradient of biomarker density.

## 3.3 Simulation Convergence and Performance Metrics

After completing the simulation, we gathered several quantitative metrics to assess the overall performance of the nanorobots, which are detailed in Table 1. These metrics collectively provide insights into both the efficiency and precision of the nanorobot, indicating how effectively it navigates toward the cancer cell while responding to biomarker cues.

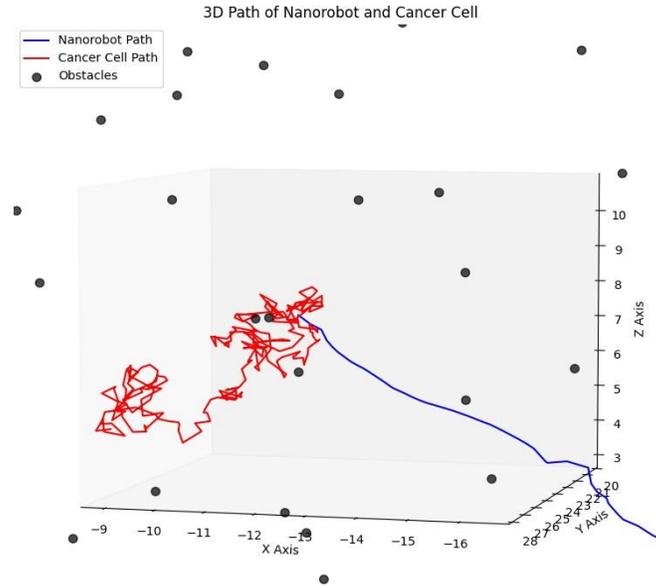

Figure 3: path of nanorobot and cancer cell.

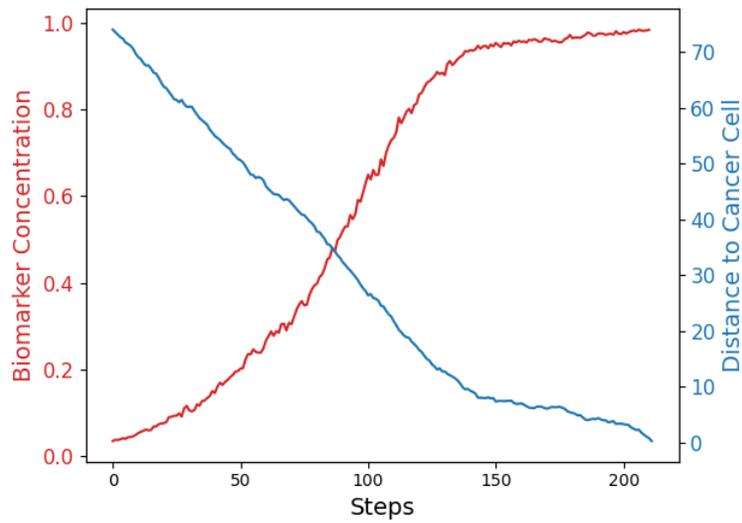

Figure 4: Biomarker concentration and distance to cancer cell over steps.

Table 1: Quantitative metrics to evaluate the nanorobot's overall performance

| Parameter | Definition | Result |
|---|---|---|
| Total Steps | The number of steps taken by the nanorobot from the start position to within a close threshold of the cancer cell. A lower step count indicates a more efficient approach. | 247 |
| Final Distance | The final Euclidean distance between the nanorobot and the cancer cell at the end of the simulation. This metric assesses the accuracy of the nanorobot's approach, ideally nearing zero. | 0.01 |
| Average Distance | The mean distance to the cancer cell throughout the simulation. This value shows how consistently the nanorobot was able to move closer to its target. | 34.66 |
| Average Biomarker Concentration | The mean concentration of biomarkers around the nanorobot during the simulation. High values indicate effective use of biomarker cues. | 0.54 |
| Simulation Duration | | 54.04 seconds |

### 3.4 Heatmap Analysis of Biomarker Distribution in the XY Plane

To further understand biomarker distribution and its effect on the nanorobot's path, we generated a heatmap to visualize biomarker concentrations. Figure 5 presents a biomarker concentration heatmap in the XY, XZ, and YZ planes, illustrating the spatial distribution of biomarkers. The XY plane offers a top-down view of biomarker concentrations, with warmer colors indicating areas of higher concentration that nanorobots may encounter. The XZ plane reveals how these concentrations vary with depth, while the YZ plane displays vertical and lateral changes. Together, these heatmaps provide valuable insights into the three-dimensional distribution of biomarkers, facilitating the effective navigation of nanorobots toward cancer cells.

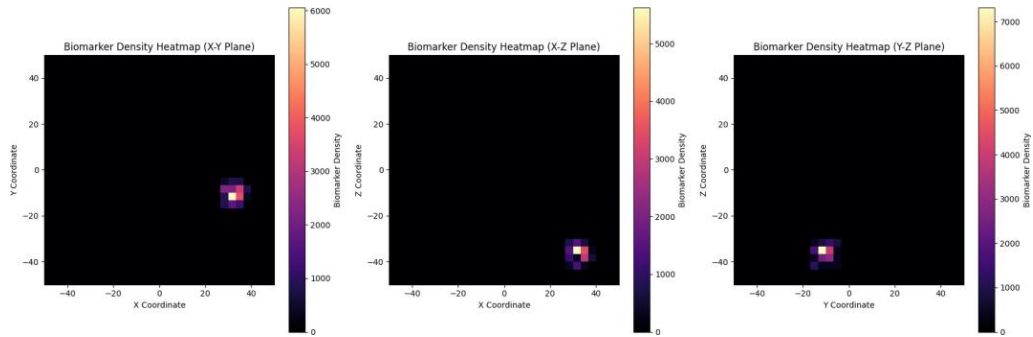

Figure 5: Biomarker concentration heatmap in the XY, XZ and YZ planes.

## 4 Discussion

The development of smart nanorobots represents a promising frontier in medicine, particularly in diagnostics and treatment. These nanorobots are capable of navigating complex biological environments and performing tasks at the cellular level, potentially revolutionizing our approach to diseases like cancer. However, the integration of this technology raises critical ethical and safety considerations. Ensuring the biocompatibility of materials used in nanorobots is essential to prevent adverse reactions in patients. Moreover, deploying autonomous systems in medical applications necessitates rigorous testing and validation to address concerns regarding their reliability and effectiveness. The scalability of manufacturing nanorobots poses another significant challenge. Current production methods may not be efficient or cost-effective enough for widespread clinical use. Overcoming these challenges will require interdisciplinary collaboration among engineers, biologists, and healthcare professionals to develop innovative yet practical solutions. As we enhance the capabilities of nanorobots, we must also consider their impact on healthcare systems and patient care. The ability to deliver targeted therapies could lead to more effective treatments with fewer side effects, but it will also necessitate changes in regulatory frameworks and healthcare practices to accommodate these new technologies.

Overall, while the potential of smart nanorobots is immense, careful consideration of ethical, technical, and practical implications will be vital for their successful integration into clinical settings. Future research should focus on several key areas to enhance the design of smart nanorobots. Integrating advanced biological sensors can improve real-time detection of cellular biomarkers, while optimizing reinforcement learning algorithms will enable nanorobots to adapt to complex environments. Expanding their applications beyond cancer detection to include drug delivery and personalized healthcare will broaden their utility. Collaboration with material scientists to develop biocompatible materials will enhance interactions with biological systems, and establishing robust communication protocols among nanorobots will improve coordination and task execution. By pursuing these strategies, we can advance the capabilities of smart nanorobots, ultimately leading to better patient outcomes and innovative therapeutic approaches. Updates to the Q-values in the Q-learning algorithm directly translate to real-world improvements in nanorobot navigation by enabling these robots to refine their movement strategies based on learned experiences. As nanorobots navigate through their environments, they encounter various conditions and challenges, such as the presence of cancer cells, biomarker gradients, and obstacles. Each time a nanorobot takes an action, it receives feedback in the form of a reward, which is used to adjust the Q-values associated with its current state and action. These updates reflect the effectiveness of the actions taken, allowing the nanorobots to learn which paths lead to successful outcomes—such as reaching cancer cells—while avoiding detrimental situations, like collisions with obstacles. Over time, the Q-values converge to represent the expected long-term rewards for different actions in specific states, leading to the development of an optimal policy. In real-world applications, this means that nanorobots can adapt their trajectories dynamically, improving their efficiency in locating and targeting cancer cells amidst the complexities of biological environments. By continually updating their Q-values, nanorobots can make informed decisions that minimize energy expenditure and maximize the likelihood of successful interactions with target cells. Consequently, this learning capability enhances their effectiveness in targeted therapies, leading to better treatment outcomes and reduced side effects in patients.

# 5 Conclusion

In this paper, we presented a simulation of nanorobots designed for the targeted detection and tracking of cancer cells using reinforcement learning techniques. The findings demonstrate the efficacy of the proposed algorithm in navigating a dynamic three-dimensional environment, wherein the cancer cells exhibit random motion. The use of biomarker gradients as navigational aids, coupled with reinforcement learning for adaptive decision-making, allows nanorobots to effectively pursue and identify their targets. The results highlight the significant potential of AI-enhanced nanorobots in biomedical applications, particularly in cancer treatment, where precise targeting is crucial for minimizing damage to healthy tissues. The simulation outcomes affirm that, with further refinement and integration of advanced algorithms, nanorobots can become vital tools in the realm of personalized medicine. Future research directions could focus on incorporating real-time biomarker sensing, enhancing obstacle avoidance strategies, and validating these simulations with empirical data from biological systems. Moreover, the application of this technology can be extended beyond cancer detection to other areas of medicine, such as drug delivery systems and treatment of various diseases, paving the way for innovative solutions in healthcare. The availability of the source code on GitHub encourages collaboration and further exploration in this promising field of study.